\documentclass[11pt,a4paper]{article}
\usepackage{times,latexsym}
\usepackage{url}
\usepackage[T1]{fontenc}

\usepackage{amsmath,bm,graphicx}
\usepackage[ruled,vlined]{algorithm2e}

\def\figref#1{Fig.~\ref{#1}}
\def\tabref#1{Table~\ref{#1}}
\def\equref#1{Eq.~\ref{#1}}

\usepackage[acceptedWithA]{tacl2018v2}

\usepackage{xspace,mfirstuc,tabulary}

\newif\iftaclinstructions
\taclinstructionsfalse 
\iftaclinstructions
\renewcommand{\confidential}{}
\renewcommand{\anonsubtext}{(No author info supplied here, for consistency with
TACL-submission anonymization requirements)}
\newcommand{\instr}
\fi

\iftaclpubformat 

\else

\fi

\title{Nested Named Entity Recognition via \\ Second-best Sequence Learning and Decoding}

\author{\bf{Takashi Shibuya}$^{\dag\ast}$ \quad \bf{Eduard Hovy}$^\dag$ \\
  $\dag$ Carnegie Mellon University, Pittsburgh, PA 15213, U.S.A. \\
  $\ast$ Sony Corporation, Tokyo 141-8610, Japan \\ 
  {\tt shibuyat@jp.sony.com} \quad {\tt hovy@cmu.edu} \\}

\date{}

\begin{document}
\maketitle
\begin{abstract}
  When an entity name contains other names within it, the identification of all combinations of names can become difficult and expensive.   
  We propose a new method to recognize not only outermost named entities but also inner nested ones.
  We design an objective function for training a neural model that treats the tag sequence for nested entities as the second best path within the span of their parent entity.
  In addition, we provide the decoding method for inference that extracts entities iteratively from outermost ones to inner ones in an outside-to-inside way.
  Our method has no additional hyperparameters to the conditional random field based model widely used for flat named entity recognition tasks.
  Experiments demonstrate that our method performs better than or at least as well as existing methods capable of handling nested entities, achieving the F1-scores of $85.82\%$, $84.34\%$, and $77.36\%$ on ACE-2004, ACE-2005, and GENIA datasets, respectively.
\end{abstract}

\section{Introduction}

Named entity recognition (NER) is the task of identifying text spans associated with proper names and classifying them according to their semantic class such as person or organization.
NER, or in general the task of recognizing entity mentions, is one of the first stages in deep language understanding, and its importance has been well recognized in the NLP community~\cite{Nadeau:2007}.

\begin{figure}[t]
\centering
\includegraphics[width=.85\hsize]{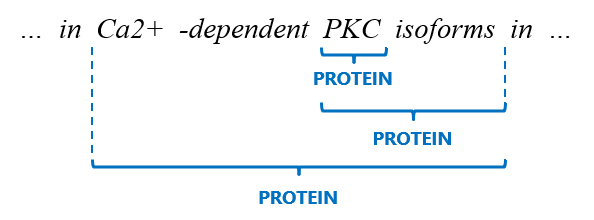}
\caption{Example of nested entities.}
\label{fig:example}
\end{figure}

One popular approach to the NER task is to regard it as a sequence labeling problem.
In this case, it is implicitly assumed that mentions are not nested in texts.
However, names often contain entities nested within themselves, as illustrated in \figref{fig:example}, which contains 3 mentions of the same type (PROTEIN) in the span ``{\it ... in Ca2+ -dependent PKC isoforms in ...}'', taken from the GENIA dataset~\cite{Kim:2003}.
Name nesting is common, especially in technical domains~\cite{alex-etal-2007-recognising,byrne2007,wang-2009-annotating}.
The assumption of no nesting leads to loss of potentially important information and may negatively impact subsequent downstream tasks.
For instance, a downstream entity linking system that relies on NER may fail to link the correct entity if the entity mention is nested.

\begin{figure*}[t]
\centering
\includegraphics[width=.85\hsize]{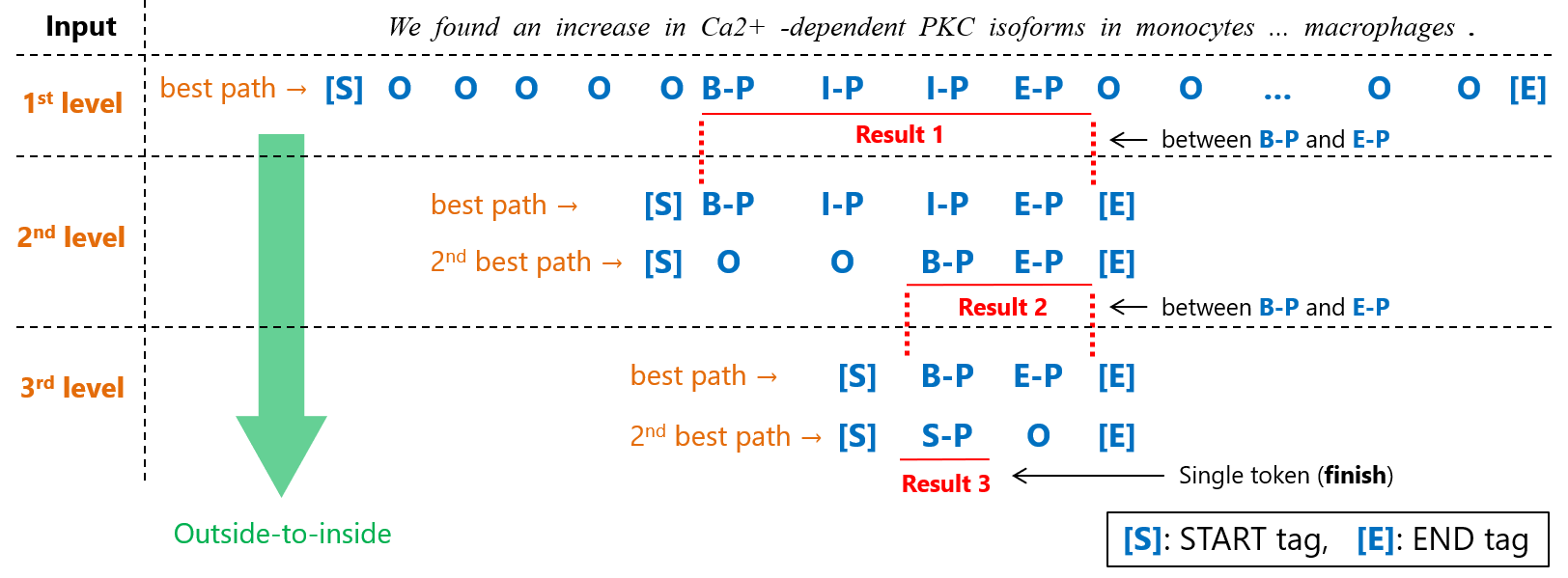}
\caption{Overview of our second-best path decoding algorithm to iteratively find nested entities.}
\label{fig:decoding}
\end{figure*}

Various approaches to recognizing nested entities have been proposed.
Many of them rely on producing and rating all possible (sub)spans, which can be computationally expensive.  
\citet{wang-lu-2018-neural} provided a hypergraph-based approach to consider all possible spans.
\citet{sohrab-miwa-2018-deep} proposed a neural exhaustive model that enumerates and classifies all possible spans.
These methods, however, achieve high performance at the cost of time complexity.
To reduce the running time, they set a threshold to discard longer entity mentions.
If the hyperparameter is set low, running time is reduced but longer mentions are missed.
In contrast, \citet{muis-lu-2017-labeling} proposed a sequence labeling approach that assigns tags to gaps between words, which efficiently handles sequences using Viterbi decoding.
However, this approach suffers from structural ambiguity issues during inference as explained by \citet{wang-lu-2018-neural}.
\citet{katiyar-cardie-2018-nested} proposed another hypergraph-based approach that learns the structure in a greedy manner.
However, their method uses an additional hyperparameter as the threshold for selecting multiple mention candidates.
This hyperparameter affects the trade-off between recall and precision. 

In this paper, we propose new learning and decoding methods to extract nested entities without any additional hyperparameters.
We summarize our contributions as follows:
\begin{itemize}
\item We describe a decoding method that iteratively recognizes entities from outermost ones to inner ones without structural ambiguity.
It recursively searches a span of each extracted entity for inner nested entities using the Viterbi algorithm.
This algorithm does not require hyperparameters for the maximal length or number of mentions considered.
\item We also provide a novel learning method that ensures the aforementioned decoding.
Models are optimized based on an objective function designed according to the decoding procedure.
\item Empirically, we demonstrate that our method performs better than or at least as well as the current state-of-the-art methods with $85.82\%$, $84.34\%$, and $77.36\%$ in F1-score on three standard datasets: ACE-2004\footnote{\url{https://catalog.ldc.upenn.edu/LDC2005T09}}, ACE-2005\footnote{\url{https://catalog.ldc.upenn.edu/LDC2006T06}}, and GENIA.
\end{itemize}

\begin{algorithm*}[t]
\small
\SetKwProg{Fn}{Function}{}{}
\SetKwFunction{Fmain}{main}
\SetKwFunction{Fnest}{detectNestedMentions}
$K$ = the set of entity types;\\
\Fn{\Fmain{$\bm{z}_i$}}{
  $M=\{\}$;\ \ \# the set of detected mentions. Each element of $M$ is a tuple ($s$, $e$, $k$) regarding a mention. \\
  \ \ \ \ \ \ \ \ \# $s$, $e$, and $k$ are the start position, the end position, and the entity type of the mention, respectively. \\
  \ForEach{$k\in K$}{
    calculate CRF scores $\bm{\Phi}$ for entity type $k$ with the score function $\phi_k\left(y_{i-1}^{\left(k\right)}, y_i^{\left(k\right)}, \bm{z}_i\right)$\;
    find the best path of the span from position $1$ to position $n$ based on the scores $\bm{\Phi}$\;
    $\Tilde{M}$ = the set of the mentions detected in the best path\;
    $M=M\cup\Tilde{M}$\;
    \ForEach{$m\in\Tilde{M}$}{
      \Fnest{$\bm{\Phi}$, $m.s$, $m.e$, $k$, $M$}\;
    }
  }
  \KwRet $M$\;
}
\Fn{\Fnest{$\bm{\Phi}$, $s$, $e$, $k$, $M$}}{
  \If{$e-s>1$}{
    find the 2nd best path of the span from position $s$ to position $e$ based on the scores $\bm{\Phi}$\;
    $\Tilde{M}$ = the set of the mentions detected in the 2nd best path\;
    $M=M\cup\Tilde{M}$\;
    \ForEach{$m\in\Tilde{M}$}{
      \Fnest{$\bm{\Phi}$, $m.s$, $m.e$, $k$, $M$}\;
    }
  }
  \KwRet\;
}
\caption{Nested NER via 2nd-best sequence decoding}\label{alg:pseudo-code}
\end{algorithm*}

\section{Method}

We propose applying conditional random field (CRF)~\cite{lafferty2001}, which is commonly used for flat NER~\cite{lample-etal-2016-neural,ma-hovy-2016-end,chiu-nichols-2016-named,reimers-gurevych-2017-reporting,strubell-etal-2017-fast,akbik-etal-2018-contextual}, to nested NER in this study.
We first explain our usage of CRF, which is the base of our decoding and training methods.
Then, we introduce our decoding and training methods.
Our decoding and training methods focus on the output layer of neural architectures and therefore can be combined with any neural model.

\subsection{Usage of CRF}
\label{ssec:usage}

Our decoding and training methods are based on two key points about our usage of CRF.
The first key point is that we prepare a separate CRF for each named entity type.
This enables our method to handle the situation where the same mention span is assigned multiple entity types.
The GENIA dataset indeed has such mention spans.
In the literature, \citet{muis-lu-2017-labeling} demonstrated that this approach of multiple CRFs would perform better on nested NER datasets and even a flat NER dataset than the standard approach of a single CRF for all entity types.
The second key point is that each element of the transition matrix of each CRF has a fixed value according to whether it corresponds to a legal transition ({\em e.g.}, {\tt B-X} to {\tt I-X} in IOBES tagging scheme, where {\tt X} is the name of entity type) or an illegal one ({\em e.g.}, {\tt O} to {\tt I-X}).
This is helpful for keeping the scores for tag sequences including outer entities higher than those of tag sequences including inner entities.

Formally, we use $\bm{Z}=\left\{\bm{z}_1, \dots , \bm{z}_n\right\}$ to represent a sequence output from the last hidden layer of a neural model, where $\bm{z}_i$ is the vector for the $i$-th word, and $n$ is the number of tokens.
$\bm{y}^{\left(k\right)}=\{y_1^{\left(k\right)}, \dots , y_n^{\left(k\right)}\}$ represents a sequence of IOBES tags of entity type $k$ for $\bm{Z}$. Here, we define the score function to be
\begin{align}
& \phi_k\left(y_{i-1}^{\left(k\right)}, y_i^{\left(k\right)}, \bm{z}_i\right)=\bm{P}_{y_i^{\left(k\right)},i}^{\left(k\right)}+\bm{A}_{y_{i-1}^{\left(k\right)}, y_i^{\left(k\right)}}^{\left(k\right)},\label{eq:score}
\end{align}
\begin{align}
& \hspace{3mm}\text{where}\ \ \ \bm{P}_{y_i^{\left(k\right)},i}^{\left(k\right)}=\bm{W}_{y_i^{\left(k\right)}}^{\left(k\right)}\cdot\bm{z}_i+\bm{b}_{y_i^{\left(k\right)}}^{\left(k\right)},\nonumber\\
& \hspace{3mm}\bm{A}_{y_{i-1}^{\left(k\right)}, y_i^{\left(k\right)}}^{\left(k\right)}=
\begin{cases}
-\infty, & \text{if $y_{i-1}^{\left(k\right)}\to y_i^{\left(k\right)}$ is illegal}, \\
0, & \text{otherwise}.
\end{cases}
\nonumber
\end{align}
$\bm{W}_{y_i^{\left(k\right)}}^{\left(k\right)}$ and $\bm{b}_{y_i^{\left(k\right)}}^{\left(k\right)}$ denote the weight matrix and the bias vector corresponding to $y_i^{\left(k\right)}$, respectively.
$\bm{A}^{\left(k\right)}$ stands for the transition matrix from the previous token to the current token, and $\bm{A}_{y_{i-1}^{\left(k\right)}, y_i^{\left(k\right)}}^{\left(k\right)}$ is the transition scores from $y_{i-1}^{\left(k\right)}$ to $y_i^{\left(k\right)}$.
$\bm{Z}$ is shared between all of the multiple CRFs as their input.

\subsection{Decoding}

We employ three strategies for decoding.
First, we consider each entity type separately using multiple CRFs in decoding, which makes it possible to handle the situation that the same mention span is assigned multiple entity types.
Second, our decoder searches nested entities in an outside-to-inside way\footnote{Our usage of {\it inside}/{\it outside} is different from the inside-outside algorithm in dynamic programming.}, which realizes efficient processing by eliminating the spans of non-entity at an early stage.
More specifically, our method recursively narrows down the spans to Viterbi-decode.
The spans to Viterbi-decode are dynamically decided according to the preceding Viterbi-decoding result.
Only the spans that have just been recognized as entity mentions are Viterbi-decoded again.
Third, we use the same scores $\phi_k\left(y_{i-1}^{\left(k\right)}, y_i^{\left(k\right)}, \bm{z}_i\right)$ of \equref{eq:score} to extract outermost entities and even inner entities without re-encoding, which makes inference more efficient and faster.
These three strategies are deployed and completed only in the output layer of neural architectures.

We describe the pseudo-code of our decoding method in Algorithm \ref{alg:pseudo-code}.
Also, we depict the overview of our decoding method with an example in \figref{fig:decoding} .
We use the term {\it level} in the sense of the depth of entity nesting.
{\tt [S]} and {\tt [E]} in \figref{fig:decoding} stand for the START and END tags respectively. 
We always attach these tags to both ends of every sequence of IOBES tags in Viterbi-decoding.

We explain the decoding procedure and mechanism in detail below.
We consider each entity type separately and iterate the same decoding process regarding distinct entity types as described in Algorithm \ref{alg:pseudo-code}.
In the decoding process for each entity type $k$, we first calculate the CRF scores $\phi_k\left(y_{i-1}^{\left(k\right)}, y_i^{\left(k\right)}, \bm{z}_i\right)$ over the entire sentence.
Next, we decode a sequence with the standard $1$-best Viterbi decoding as with the conventional linear-chain CRF.
``{\it Ca2+ -dependent PKC isoforms}'' is extracted at the 1st level with regard to the example of \figref{fig:decoding}.

Then, we start our recursive decoding to extract nested entities within previously extracted entity spans by finding the 2nd best path.
In \figref{fig:decoding}, the span ``{\it Ca2+ -dependent PKC isoforms}'' is processed at the 2nd level.
Here, if we search for the best path within each span, the same tag sequence will be obtained, even though the processed span is different.
This is because we continue using the same scores $\phi_k\left(y_{i-1}^{\left(k\right)}, y_i^{\left(k\right)}, \bm{z}_i\right)$ and because all the values of $\bm{A}^{\left(k\right)}$ corresponding to legal transitions are equal to $0$.
Regarding the example of \figref{fig:decoding}, the score of the transition from {\tt [S]} to {\tt B-P} at the 2nd level is equal to the score of the transition from {\tt O} to {\tt B-P} at the 1st level.
This is true for the transition from {\tt E-P} to {\tt [E]} at the 2nd level and the one from {\tt E-P} to {\tt O} at the 1st level.
The best path between the {\tt [S]} and {\tt [E]} tags is identical to the best path between the two {\tt O} tags under our restriction about the transition matrix of CRF.
Therefore, we search for the 2nd best path within the span by utilizing the $N$-best Viterbi A* algorithm~\cite{Seshadri:1994,huang-etal-2012-iterative}.\footnote{Without our restriction about the transition matrix of CRF, we would have to watch both the best path and the 2nd best path. Besides, if a single CRF was used for all entity types, the decoder could not always narrow down spans with the 2nd best path. The 2nd best path in a single CRF could result in the same span tagged a different entity type. We would have to watch lower-ranked paths.}
Note that our situation is different from normal situations where $N$-best decoding is needed.
We already know the best path within the span and want to find only the 2nd best path.
Thus, we can extract nested entities by finding the 2nd best path within each extracted entity.
Regarding the example of \figref{fig:decoding}, ``{\it PKC isoforms}'' is extracted from the span ``{\it Ca2+ -dependent PKC isoforms}'' at the 2nd level.

We continue this recursive decoding until no multi-token entities are detected within a span.
In \figref{fig:decoding}, the span ``{\it PKC isoforms}'' is processed at the 3rd level.
At the 3rd or deeper levels, the tag sequence of its grandparent level is no longer either the best path or the 2nd best path because the start or end position of the current span is in the middle of the entity mention span at the grandparent level.
As for the example shown in \figref{fig:decoding}, the word ``{\it PKC}'' is tagged {\tt I-P} at the 1st level, and the transition from {\tt [S]} to {\tt I-P} is illegal.
The scores of the paths that includes illegal transitions cannot be larger than those of the paths that consist of only legal transitions because the elements of the transition matrix $\bm{A}^{\left(k\right)}$ corresponding to illegal transitions are set to $-\infty$.
That is why at all levels below the 1st level we only need to find the 2nd best path.

This recursive processing is stopped when no entities are predicted or when only single-token entities are detected within a span.\footnote{We do not need to recursively decode the span of each extracted single-token entity because a single-token entity cannot contain another entity of the same entity type.}
In \figref{fig:decoding}, the span ``{\it PKC}'' is not processed any more because it is a single-token entity.

Only one nested entity is extracted within each decoded span in \figref{fig:decoding}, but there can be cases where multiple multi-token entities are detected within a decoded span.
In such cases, our algorithm Viterbi-decodes each of their spans in the way of the depth-first search algorithm.
The aforementioned processing is executed on all entity types, and all detected entities are returned as an output result.

\subsection{Training}

To extract entities from outside to inside successfully, a model has to be trained in a way that the scores for the paths including outer entities will be higher than those for the paths including inner entities.
We propose a new objective function to achieve this requirement.

We maximize the log-likelihood of the correct tag sequence as with the conventional CRF-based model.
Considering that our model has a separate CRF for each entity type, the log-likelihood for one training data, $\mathcal{L}\left(\bm{\theta}\right)$, is as follows:
\begin{equation}
\mathcal{L}\left(\bm{\theta}\right)=\sum_{k}\log p\left(\bm{Y}^{\left(k\right)}|\bm{Z}; \bm{\theta}\right),\label{eq:log-likelihood}
\end{equation}
where $\bm{\theta}$ is the set of parameters of a neural model, and $\bm{Y}^{\left(k\right)}$ denotes the collection of the gold IOBES tags for all levels regarding the entity type $k$.
As we mentioned in Section \ref{ssec:usage}, $\bm{Z}$ is a sequence output from the last hidden layer of a neural model and is shared between all of the multiple CRFs.
Therefore, $\bm{\theta}$ is updated through a backpropagation process so that $\bm{Z}$ can represent information about all entity types.

In the following, we decompose the log-likelihood for all levels into the ones for each level.
Let $s_{l,j}^{\left(k\right)}$ and $e_{l,j}^{\left(k\right)}$ denote the start and end positions of the $j$-th span at the $l$-th level.
With regard to the 1st level, $s_{1,1}^{\left(k\right)}=1$ and $e_{1,1}^{\left(k\right)}=n$ because we consider the whole span of a sentence.
The spans considered at each deeper level, $l > 1$, are determined according to the spans of multi-token entities at its immediate parent level.
As for the example of \figref{fig:decoding}, only the span of ``{\it Ca2+ -dependent PKC isoforms}'' is considered at the 2nd level.
Here, the log-likelihood for each entity type can be expressed as follows:
\begin{align}
& \log p\left(\bm{Y}^{\left(k\right)}|\bm{Z}; \bm{\theta}\right)=L_{\text{1st}}\left(y_{1,1}^{\left(k\right)}, \dots , y_{1,n}^{\left(k\right)}|\bm{Z}; \bm{\theta}\right)\nonumber\\
& \hspace{10mm}+\sum_{l>1}\sum_jL_{\text{2nd}}\left(y_{l,s_{l,j}^{\left(k\right)}}^{\left(k\right)}, \dots , y_{l,e_{l,j}^{\left(k\right)}}^{\left(k\right)}|\bm{Z}; \bm{\theta}\right),\label{eq:log-likelihood-each}
\end{align}
where $L_{\text{1st}}\left(\dots\right)$ and $L_{\text{2nd}}\left(\dots\right)$ are the log-likelihoods of the (1st) best and 2nd best paths for each span, respectively.
$y_{l,i}^{\left(k\right)}$ denotes the correct IOBES tag of the position $i$ of the $l$-th level of the entity type $k$.

\begin{algorithm}[t]
\small
$C=\{\text{\tt B-X}, \text{\tt I-X}, \text{\tt E-X}, \text{\tt S-X}, \text{\tt O}\}$\;
$s=1$;\ \ \# the start position \\
$e=n$;\ \ \# the end position \\
\ForEach{$c\in C$}{
  $\bm{\alpha}\left(c\right)=\bm{P}_{c,s}^{\left(k\right)}+\bm{A}_{{\tt [S]},c}^{\left(k\right)}$\;
}
\For{$i=s+1$; $i\leq e$; $i++$}{
  \ForEach{$c\in C$}{
    \ForEach{$c'\in C$}{
      $\bm{\alpha}_{c}\left(c'\right)=\bm{\alpha}\left(c'\right)+\bm{P}_{c,i}^{\left(k\right)}+\bm{A}_{c',c}^{\left(k\right)}$\;
    }
  }
  \ForEach{$c\in C$}{
    $\bm{\alpha}\left(c\right)=\text{LogSumExp}\left(\bm{\alpha}_{c}\right)$\;
  }
}
\ForEach{$c\in C$}{
  $\bm{\alpha}\left(c\right)+=\bm{A}_{c,{\tt [E]}}^{\left(k\right)}$\;
}
\KwRet $\text{LogSumExp}\left(\bm{\alpha}\right)$\;
\caption{LogSumExp of the scores of all possible paths}\label{alg:1st}
\end{algorithm}

\noindent {\bf Best path.} $L_{\text{1st}}\left(\dots\right)$ can be calculated in the same manner as the conventional linear-chain CRF:
\begin{align}
& L_{\text{1st}}\left(y_{1,1}^{\left(k\right)}, \dots , y_{1,n}^{\left(k\right)}|\bm{Z}; \bm{\theta}\right)=\nonumber\\
& \hspace{4mm}\psi_{1:n}^{\left(k\right)}\left(\bm{y}_{1,1}^{\left(k\right)}, \bm{Z}\right)-\log\sum_{\bm{y'}\in\mathcal{Y}_{1:n}^{\left(k\right)}}\exp\psi_{1:n}^{\left(k\right)}\left(\bm{y'}, \bm{Z}\right),\label{eq:log-likelihood-1st}\\
& \hspace{5mm}\text{where}\ \ \ \psi_{s:e}^{\left(k\right)}\left(\bm{y}, \bm{Z}\right)=\nonumber\\
& \hspace{26mm}\sum_{i=s}^{e}\phi_k\left(y_{i-1}, y_i, \bm{z}_i\right)+\bm{A}_{y_{e}, y_{e+1}}^{\left(k\right)},\nonumber\\
& \hspace{17mm} y_{s-1}=\text{\tt [S]},\ y_{e+1}=\text{\tt [E]}.\nonumber
\end{align}
$\mathcal{Y}_{s:e}^{\left(k\right)}$ denotes the set of all possible tag sequences from position $s$ to position $e$ of the entity type $k$.
The first term of \equref{eq:log-likelihood-1st} is the score of the gold tag sequence, and the second term is the logarithm of the summation of the exponential scores of all possible tag sequences.
It is well known that the second term of \equref{eq:log-likelihood-1st} can be efficiently calculated by the algorithm shown in Algorithm \ref{alg:1st}.

\begin{figure}[t]
\centering
\includegraphics[width=.58\hsize]{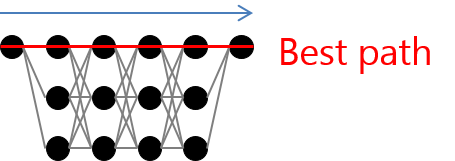}
\caption{Lattice and best path.}
\label{fig:best-path}
\end{figure}

\begin{figure*}[t]
\centering
\includegraphics[width=.85\hsize]{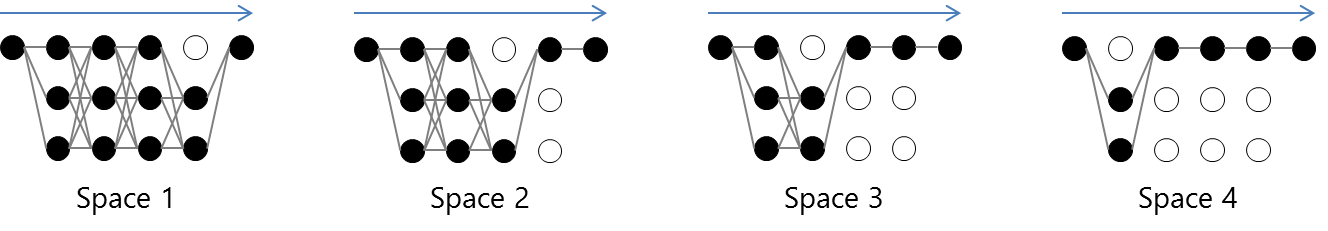}
\caption{Divided search spaces.}
\label{fig:devided-paths}
\end{figure*}

\begin{figure}[t]
\centering
\includegraphics[width=.95\hsize]{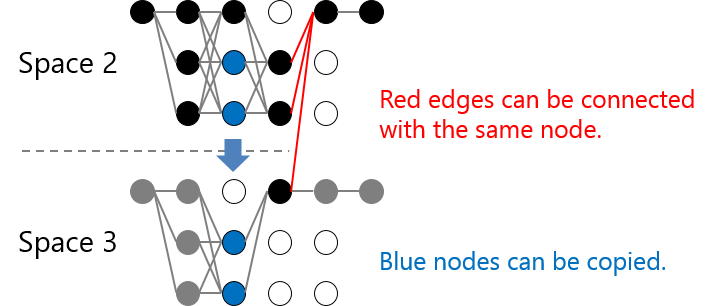}
\caption{Merge of search spaces.}
\label{fig:merged-paths}
\end{figure}

\begin{algorithm}[t]
\small
$C=\{\text{\tt B-X}, \text{\tt I-X}, \text{\tt E-X}, \text{\tt S-X}, \text{\tt O}\}$\;
$s=s_{l,j}^{\left(k\right)}$;\ \ \# the start position \\
$e=e_{l,j}^{\left(k\right)}$;\ \ \# the end position \\
$\bm{c}_1\left(s\right)=\text{\tt B-X}$;\ \ \# the best path \\
\For{$i=s+1$; $i\leq e-1$; $i++$}{
  $\bm{c}_1\left(i\right)=\text{\tt I-X}$\;
}
$\bm{c}_1\left(e\right)=\text{\tt E-X}$\;
\ForEach{$c\in C$}{
  $\bm{\alpha}\left(c\right)=\bm{P}_{c,s}^{\left(k\right)}+\bm{A}_{{\tt [S]},c}^{\left(k\right)}$\;
}
$\beta=-\infty$\;
\For{$i=s+1$; $i\leq e$; $i++$}{
  \ForEach{$c\in C$}{
    \ForEach{$c'\in C$}{
      $\bm{\alpha}_{c}\left(c'\right)=\bm{\alpha}\left(c'\right)+\bm{P}_{c,i}^{\left(k\right)}+\bm{A}_{c',c}^{\left(k\right)}$\;
    }
    \If{$c==\bm{c}_1(i)$}{
      \ForEach{$c'\in C\backslash\{\bm{c}_1(i-1)\}$}{
        $\bm{\beta}_{c}\left(c'\right)=\bm{\alpha}_{c}\left(c'\right)$\;
      }
      $\bm{\beta}_{c}\left(\bm{c}_1(i-1)\right)=\beta+\bm{P}_{c,i}^{\left(k\right)}+\bm{A}_{\bm{c}_1(i-1),c}^{\left(k\right)}$\;
    }
  }
  \ForEach{$c\in C$}{
    $\bm{\alpha}\left(c\right)=\text{LogSumExp}\left(\bm{\alpha}_{c}\right)$\;
  }
  $\beta=\text{LogSumExp}\left(\bm{\beta}_{c}\right)$\;
}
\ForEach{$c\in C\backslash\{\bm{c}_1(e)\}$}{
  $\bm{\alpha}\left(c\right)+=\bm{A}_{c,{\tt [E]}}^{\left(k\right)}$\;
}
$\bm{\alpha}\left(\bm{c}_1(e)\right)=\beta+\bm{A}_{\text{\tt E-X},{\tt [E]}}^{\left(k\right)}$\;
\KwRet $\text{LogSumExp}\left(\bm{\alpha}\right)$\;
\caption{LogSumExp of the scores of all possible paths except the best path}\label{alg:2nd}
\end{algorithm}

\noindent {\bf 2nd best path.} $L_{\text{2nd}}\left(\dots\right)$ given the best path can be calculated by excluding the best path from all possible paths.
This concept is also adopted by ListNet~\cite{cao2007}, which is used for ranking tasks such as document retrieval or recommendation.
$L_{\text{2nd}}\left(\dots\right)$ can be expressed by the following equation:
\begin{align}
& L_{\text{2nd}}\left(y_{l,s_{l,j}^{\left(k\right)}}^{\left(k\right)}, \dots , y_{l,e_{l,j}^{\left(k\right)}}^{\left(k\right)}|\bm{Z}; \bm{\theta}\right)=\nonumber\\
& \hspace{5mm}\psi_{s_{l,j}^{\left(k\right)}:e_{l,j}^{\left(k\right)}}^{\left(k\right)}\left(\bm{y}_{l,j}^{\left(k\right)}, \bm{Z}\right)\nonumber\\
& \hspace{13mm}-\log\sum_{\bm{y'}\in\Tilde{\mathcal{Y}}_{s_{l,j}^{\left(k\right)}:e_{l,j}^{\left(k\right)}}^{\left(k\right)}}\exp\psi_{s_{l,j}^{\left(k\right)}:e_{l,j}^{\left(k\right)}}^{\left(k\right)}\left(\bm{y'}, \bm{Z}\right),\label{eq:log-likelihood-2nd}
\end{align}
where $\Tilde{\mathcal{Y}}_{s:e}^{\left(k\right)}$ denotes the set of all possible tag sequences except the best path within the span from position $s$ to position $e$ of the entity type $k$.

However, to the best of our knowledge, the way of efficiently computing the second term of \equref{eq:log-likelihood-2nd} has not been proposed yet in the literature.
Simply subtracting the exponential score of the best path from the summation of the exponential scores of all possible paths causes underflow, overflow, or loss of significant digits.
We introduce the way of accurately computing it with the same time complexity as Algorithm \ref{alg:1st} for \equref{eq:log-likelihood-1st}.
For explanation, we use the simplified example of the lattice depicted in \figref{fig:best-path}, in which the span length is $4$ and the number of states is $3$.
The special nodes for start and end states are attached to the both ends of the span.
There are $81(=3^4)$ paths in this lattice.
We assume that the path that consists of top nodes of all time steps are the best path as shown in \figref{fig:best-path}.
No generality is lost by making this assumption.
To calculate the second term of \equref{eq:log-likelihood-2nd}, we have to consider the exponential scores for all the possible paths except the best path, $80(=81-1)$ paths.

We first give a way of thinking, which is not our algorithm itself but helpful to understand it.
In the example, we can further group these 80 paths according to the steps where the best path is not taken.
In this way, we have $4$ spaces in total as illustrated in \figref{fig:devided-paths}.
In Space $1$, the top node of time step $4$ is excluded from consideration.
$54(=3^3\times2)$ paths are taken into account here.
Since this space covers all paths that do not go through the top node of time step $4$, we only have to consider the paths that go through this node in other spaces.
In Space $2$, this node is always passed through, and instead the top node of time step $3$ is excluded. 
$18(=3^2\times2)$ paths are considered in this space.
Similarly, $6(=3^1\times2)$ paths and $2(=3^0\times2)$ paths are taken into consideration in Space $3$ and Space $4$, respectively.
Thus, we can consider all the possible paths except the best path, $80(=54+18+6+2)$ paths.
However, this is not our algorithm itself as we mentioned.

We introduce two tricks for making the calculation more efficient.
We explain them with \figref{fig:merged-paths}, in which Spaces $2$ and $3$ are picked up.
The first trick is that the separated two spaces can be merged at time step $4$ because the paths later than time step $3$ are identical.
When we reach time step $4$ in the forward iteration in each of the two spaces, we can merge them using the calculation results at time step $3$, as shown with the red edges in \figref{fig:merged-paths}.
The second trick is that the blue nodes in \figref{fig:merged-paths} can be copied from Space $2$ to Space $3$ at time step $2$ since the considered paths until that time step are also the same.
These two tricks can be applied to other pairs of two adjacent spaces, which relieves the need to separately calculate the summation of the exponential scores for each space.
Therefore, the second term of \equref{eq:log-likelihood-2nd} can be calculated as shown in Algorithm \ref{alg:2nd}.

Thus, we can train a model using the objective function of Eqs.~\ref{eq:log-likelihood}, \ref{eq:log-likelihood-each}, \ref{eq:log-likelihood-1st}, and \ref{eq:log-likelihood-2nd}.

\subsection{Characteristics}

\noindent {\bf Time complexity.} 
Regarding the time complexity of decoder, the worst case for our method is when our decoder narrows down the spans one by one, from $n$ tokens (a whole sentence) to $2$ tokens.
The time complexity for the worst case is therefore $\mathcal{O}\left(n + \dots + 2\right) = \mathcal{O}\left(n^2\right)$ for each entity type, $\mathcal{O}\left(mn^2\right)$ in total, where $m$ denotes the number of entity types.
However, this rarely happens.
The ideal average processing time in the case where our decoding method narrows down spans successfully according to gold labels is $\mathcal{O}\left(dmn\right)$, where $d$ is the average number of gold IOBES tags of each entity type assigned to a word.
The average numbers calculated from the gold labels of ACE-2004, ACE-2005, and GENIA are $1.06$, $1.06$, and $1.05$, respectively.

\noindent {\bf Usability.}
Some existing methods have hyperparameters, such as the maximal length of considered entities or the threshold that affects the number of detected entities, beyond those of the conventional CRF-based model used for flat NER tasks.
These hyperparameters must be tuned depending on datasets. 
On the other hand, our method does not have such hyperparameters and is easy to use from this viewpoint.
In addition, our method focuses on the output layer of neural architectures; therefore our method can be combined with any neural model.

We verify the empirical performances of our methods in the successive sections.

\section{Experimental Settings}

\subsection{Datasets}

\begin{table*}[t!]
\centering
\small
\resizebox{\textwidth}{!}{
\begin{tabular}{l|rr|rr|rr|rr|rr|rr}
  & \multicolumn{6}{c|}{\bf ACE-2005} & \multicolumn{6}{c}{\bf GENIA} \\
  &
  Train & ($\%$) & Dev & ($\%$) & Test & ($\%$) &
  Train & ($\%$) & Dev & ($\%$) & Test & ($\%$) \\
  \hline
  \# documents & 370 & & 43 & & 51 & & - & & - & & - & \\ \hline
  \# sentences &
  (7,285) & & (968) & & (1,058) & & 15,022 & & 1,669 & & 1,855 & \\ \hline
  \# mentions &
  24,827 & & 3,234 & & 3,041 & & 47,027 & & 4,469 & & 5,600 & \\
  \ \ \ - 1st level &
  21,966 & (88) & 2,900 & (90) & 2,686 & (88) & 44,611 & (95) & 4,239 & (95) & 5,273 & (94) \\
  \ \ \ - 2nd level &
  2,635 & (11) & 316 & (10) & 323 & (11) & 2393 & (5) & 230 & (5) & 327 & (6) \\
  \ \ \ - 3rd level &
  215 & (1) & 18 & (1) & 30 & (1) & 23 & (0) & 0 & (0) & 0 & (0) \\
  \ \ \ - 4th level &
  9 & (0) & 0 & (0) & 2 & (0) & 0 & (0) & 0 & (0) & 0 & (0) \\ \hline
  \# labels per token ($d$) & 1.06 & & 1.05 & & 1.05 & & 1.05 & & 1.05 & & 1.05 & \\
\end{tabular}
}
\caption{Statistics of the datasets used in the experiments. Note that in ACE-2005, sentences are not originally split. We report the numbers of sentences based on the preprocessing with the Stanford CoreNLP~\cite{manning-etal-2014-stanford}.}\label{tab:stats-datasets}
\end{table*}

We perform nested entity extraction experiments intensively on ACE-2005~\cite{doddington-etal-2004-automatic} and GENIA~\cite{Kim:2003}.
For ACE-2005, we use the same splits of documents as \citet{lu-roth-2015-joint}, published on their website\footnote{\url{http://www.statnlp.org/research/ie}}.
For GENIA, we use GENIAcorpus3.02p\footnote{\url{http://www.geniaproject.org/genia-corpus/pos-annotation}} in which sentences are already tokenized~\cite{tateisi-tsujii-2004-part}.
Following previous works~\cite{finkel-manning-2009-nested,lu-roth-2015-joint}, we first split the last $10\%$ of sentences as the test set.
Next, we use the first $81\%$ and the subsequent $9\%$ for training and development sets, respectively.
We make the same modifications as described by \citet{finkel-manning-2009-nested} by collapsing all DNA, RNA, and protein subtypes into DNA, RNA, and protein, keeping cell line and cell type, and removing other entity types, resulting in 5 entity types.
The statistics of each dataset are shown in \tabref{tab:stats-datasets}.

\subsection{Model and Training}

\begin{table}[t!]
\centering
\small
\begin{tabular}{l|ccc}
  {\bf Hyperparameter} & {\bf Value} \\
  \hline
  word dropout rate & $0.05$ \\
  character embedding dimension & $128$ \\
  CNN window size & $3$ \\
  CNN filter number & $256$ \\
  \hline
  batch size & $32$ \\
  LSTM hidden size & $256$ \\
  LSTM dropout rate & $0.2$ (w/o BERT) \\
   & $0.5$ (w/ BERT) \\
  gradient clipping & $5.0$ \\
\end{tabular}
\caption{Hyperparameters in our experiments.}\label{tab:hyperparameters}
\end{table}

In this study, we adopt as baseline a BiLSTM-CRF model, which is widely used for NER tasks~\cite{lample-etal-2016-neural,ma-hovy-2016-end,chiu-nichols-2016-named,reimers-gurevych-2017-reporting}.
We apply our usage of CRF to this baseline.
We prepare three types of models for fair comparisons with existing methods.
The first one is the model to which is fed conventional word embeddings and CNN-based character-level representation~\cite{ma-hovy-2016-end,chiu-nichols-2016-named,reimers-gurevych-2017-reporting}.\footnote{\url{https://github.com/yahshibu/nested-ner-tacl2020}}
We initialize word embeddings with the pretrained embeddings GloVe~\cite{pennington-etal-2014-glove} of dimension 100 in ACE-2005.
For GENIA, we adopt the pretrained embeddings trained on MEDLINE abstracts~\cite{chiu-etal-2016-train} instead.
The initialized word embeddings are fixed during training.
The vectors of the word embeddings and the character-level representation are concatenated and then input into the BiLSTM layer.
The second model is the model combined with the pretrained BERT model~\cite{devlin-etal-2019-bert}.\footnote{\url{https://github.com/yahshibu/nested-ner-tacl2020-transformers}}
We use the uncased version of BERT large model as a contextual word embeddings generator without fine-tuning and stack the BiLSTM layers on top of the BERT model.
The third model is the BiLSTM-CRF model to which is fed word embeddings, character-level representation, BERT embeddings, and FLAIR embeddings~\cite{akbik-etal-2018-contextual} using FLAIR framework~\cite{akbik-etal-2019-flair}.\footnote{\url{https://github.com/yahshibu/nested-ner-tacl2020-flair}}
All our models have 2 BiLSTM hidden layers, and the dimensionality of each hidden unit is 256 in all our experiments.
\tabref{tab:hyperparameters} lists the hyperparameters used for our experimental evaluations.
We adopt AdaBound~\cite{luo2019} as an optimizer.
Early stopping is used based on the performance of development set.
We repeat the experiment 5 times with different random seeds and report average and standard deviation of F1-scores on a test set as the final performance.

\section{Experimental Results}

\subsection{Comparison with Existing Methods}

\begin{table*}[t!]
\centering
\small
\resizebox{\textwidth}{!}{
\begin{tabular}{l|lll|lll}
   & \multicolumn{3}{c|}{\bf ACE-2005} & \multicolumn{3}{c}{\bf GENIA} \\
  {\bf Method} &
  Precision ($\%$) & Recall ($\%$) & F1 ($\%$) &
  Precision ($\%$) & Recall ($\%$) & F1 ($\%$) \\
  \hline
  \citet{katiyar-cardie-2018-nested} &
  $70.6$ & $70.4$ & $70.5$ &
  $79.8$ & $68.2$ & $73.6$ \\
  \citet{ju-etal-2018-neural}\footnotemark &
  $74.2$ & $70.3$ & $72.2$ &
  $78.5$ & $71.3$ & $74.7$ \\
  \citet{wang-etal-2018-neural-transition}$^\dag$\footnotemark &
  $74.5$ & $71.5$ & $73.0$ &
  $78.0$ & $70.2$ & $73.9$ \\
  \citet{wang-lu-2018-neural}$^\dag$&
  $76.8$ & $72.3$ & $74.5$ &
  $77.0$ & $73.3$ & $75.1$ \\
  \citet{sohrab-miwa-2018-deep} &
  - & - & - &
  $93.2$ & $64.0$ & $77.1$ \\
  \citet{zheng-etal-2019-boundary} &
  - & - & - &
  $75.9$ & $73.6$ & $74.7$ \\
  \citet{fisher-vlachos-2019-merge} &
  $75.1$ & $74.1$ & $74.6$ &
  - & - & - \\
  \citet{lin-etal-2019-sequence}$^\dag$&
  $76.2$ & $73.6$ & $74.9$ &
  $75.8$ & $73.9$ & $74.8$ \\
  \citet{strakova-etal-2019-neural}$^\dag$\footnotemark &
  $76.35$ & $74.39$ & $75.36$ &
  $79.60$ & $73.53$ & $76.44$ \\
  {\bf This work} &
  $78.27 \pm 0.81$ & $75.44 \pm 0.37$ & $\bm{76.83} \pm 0.36$ &
  $78.70 \pm 0.69$ & $75.74 \pm 0.64$ & $\bm{77.19} \pm 0.10$ \\ \hline
  \citet{fisher-vlachos-2019-merge} [BERT] &
  $82.7$ & $82.1$ & $82.4$ &
  - & - & - \\
  \citet{strakova-etal-2019-neural} [BERT]$^\dag$ &
  $82.58$ & $84.29$ & $83.42$ &
  $79.92$ & $76.55$ & $\bm{78.20}$ \\
  {\bf This work} [BERT] &
  $83.30 \pm 0.22$ & $84.69 \pm 0.37$ & $\bm{83.99} \pm 0.27$ &
  $77.46 \pm 0.65$ & $76.65 \pm 0.58$ & $77.05 \pm 0.12$ \\ \hline
  \citet{strakova-etal-2019-neural} [BERT+FLAIR]$^\dag$ &
  $83.48$ & $85.21$ & $\bm{84.33}$ &
  $80.11$ & $76.60$ & $\bm{78.31}$ \\
  {\bf This work} [BERT+FLAIR] &
  $83.83 \pm 0.39$ & $84.87 \pm 0.09$ & $\bm{84.34} \pm 0.20$ &
  $77.81 \pm 0.69$ & $76.94 \pm 1.12$ & $77.36 \pm 0.26$ \\
\end{tabular}
}
\caption{Main results. We group methods into three types. The first group consists of the methods that do not use any contextual word embeddings. The second group consists of the methods that use BERT but do not use any other contextual word embeddings. The third group consists of the methods that use both BERT and FLAIR. ``$\dag$'' indicates the methods using POS tags.}\label{tab:main-results}
\end{table*}

\tabref{tab:main-results} presents comparisons of our model with existing methods.
Note that some existing methods use embeddings of POS tags as an additional input feature whereas our method does not.
Our method outperforms the existing methods with $76.83\%$ and $77.19\%$ in terms of F1-score when using only word embeddings and character-level representation.
Especially, our method brings much higher recall values than the other methods.
The recall scores are improved by $3.1\%$ and $2.4\%$ on ACE-2005 and GENIA datasets, respectively.
These results demonstrate that our training and decoding algorithms are quite effective for extracting nested entities.
Moreover, when we use BERT and FLAIR as contextual word embeddings, we achieve an F1-score of $83.99\%$ with BERT and $84.34\%$ with BERT and FLAIR on ACE-2005.
On the other hand, BERT does not perform well on GENIA.
We assume that this is because the domain of GENIA is quite different from that of the corpus used for training the BERT model.
Regardless, it is demonstrated that our method performs better than or at least as well as existing methods.

\addtocounter{footnote}{-2}
\footnotetext{Note that in ACE-2005, \citet{ju-etal-2018-neural} did their experiments with a different split from \citet{lu-roth-2015-joint} that we follow.}
\stepcounter{footnote}
\footnotetext{\citet{wang-etal-2018-neural-transition} did not report precision and recall scores. Instead of \citet{wang-etal-2018-neural-transition}, \citet{wang-lu-2018-neural} reported the scores for the model of \citet{wang-etal-2018-neural-transition}.}
\stepcounter{footnote}
\footnotetext{\citet{strakova-etal-2019-neural} did not report precision and recall scores in their paper. We asked the authors the scores, and they let us know.}

\subsection{Ablation Study}
\label{ssec:ablation}

\begin{table*}[t!]
\centering
\small
\resizebox{0.9\textwidth}{!}{
\begin{tabular}{l|lll|lll}
   & \multicolumn{3}{c|}{\bf ACE-2005} & \multicolumn{3}{c}{\bf GENIA} \\
   &
  Precision ($\%$) & Recall ($\%$) & F1 ($\%$) &
  Precision ($\%$) & Recall ($\%$) & F1 ($\%$) \\
  \hline
  {\bf This work} &
  $78.27 \pm 0.81$ & $75.44 \pm 0.37$ & $76.83 \pm 0.36$ &
  $78.70 \pm 0.69$ & $75.74 \pm 0.64$ & $77.19 \pm 0.10$ \\
  \ \ \ $-$ L &
  $60.89 \pm 1.30$ & $75.38 \pm 1.27$ & $67.34 \pm 0.37$ &
  $70.72 \pm 0.39$ & $79.20 \pm 1.27$ & $74.71 \pm 0.18$ \\
  \ \ \ $-$ L\&D &
  $77.77 \pm 0.31$ & $67.42 \pm 0.29$ & $72.22 \pm 0.13$ &
  $79.70 \pm 0.56$ & $73.41 \pm 0.35$ & $76.43 \pm 0.28$ \\
\end{tabular}
}
\caption{Results when ablating away the learning (L) and decoding (D) components of our proposed method.}\label{tab:ablation}
\end{table*}

We conduct an ablation study to verify the effectiveness of our learning and decoding methods.
We first replace our objective function for training with the standard objective function of the liner-chain CRF.
The methods for decoding $N$-best paths have been well studied because such algorithms have been required in many domains~\cite{soong-huang-1990-tree,kaji-etal-2010-efficient,huang-etal-2012-iterative}.
However, we hypothesize that our learning method, as well as our decoding method, helps to improve performance.
That is why we first remove only our learning method.
Then, we also replace our decoding algorithm with the standard decoding algorithm of the linear-chain CRF.
It is equivalent to preparing the conventional CRF for each entity type separately.

The results are shown in \tabref{tab:ablation}.
They demonstrate that introducing only our decoding algorithm surely brings high recall scores but hurts precision.
This suggests that our learning method should be necessary for achieving high precision. 
Besides, removing the decoding algorithm results in lower recall.
This is natural because it does not intend to find any nested entity after extracting outermost entities.
Thus, it is demonstrated that both our learning and decoding algorithms contribute much to good performance.

\subsection{Analysis of Behavior}

\begin{table}[t!]
\centering
\small
\resizebox{\columnwidth}{!}{
\begin{tabular}{c|cr|cr}
   & \multicolumn{2}{c|}{\bf ACE-2005} & \multicolumn{2}{c}{\bf GENIA} \\
  {\bf Level} & Recall ($\%$) & Num. & Rcall ($\%$) & Num. \\ \hline
  1st & $76.10 \pm 0.50$ & 2,686 & $77.92 \pm 0.72$ & 5,273 \\
  2nd & $71.70 \pm 0.70$ & 323 & $40.61 \pm 1.74$ & 327 \\
  3rd & $58.00 \pm 5.42$ & 30 & - & 0 \\
  4th & $50.00 \pm 0.00$ & 2 & - & 0 \\
\end{tabular}
}
\caption{Recall scores for gold annotations of each level.}\label{tab:level-wise-recall}
\end{table}

\begin{table}[t!]
\centering
\small
\resizebox{\columnwidth}{!}{
\begin{tabular}{c|cr|cr}
   & \multicolumn{2}{c}{\bf ACE-2005} & \multicolumn{2}{c}{\bf GENIA} \\
  {\bf Level} & Precision ($\%$) & Num. & Precision ($\%$) & Num. \\ \hline
  1st & $80.36$ & 2,500 & $80.29$ & 5,038 \\
  2nd & $72.35$ & 311 & $57.06$ & 326 \\
  3rd & $79.07$ & 43 & $66.67$ & 3 \\
  4th & $66.67$ & 9 & - & 0 \\
  5th & $83.33$ & 6 & - & 0 \\
\end{tabular}
}
\caption{Precision scores for predictions of each level of one trial.}\label{tab:level-wise-precision}
\end{table}

To further understand how our method handles nested entities, we investigate the performances for entities of each level.
\tabref{tab:level-wise-recall} shows the recall scores for gold entities of each level when using conventional word embeddings.
Among all levels, our model results in the best performance at the 1st level that consists of only gold outermost entities.
The deeper a level, the lower recall scores.
On the other hard, \tabref{tab:level-wise-precision} shows the precision scores for predicted entities in each level of one trial on each dataset.
Because the number of levels in the predictions vary between trials, taking macro average of precision scores over multiple trials is not representative.
Therefore, we show only the precision scores from one trial in \tabref{tab:level-wise-precision}.
The precision score for the 5th level on ACE-2005 is as high as or higher than those of other levels.
Precision scores are less dependent on level.
This tendency is also shown in other trials.

In addition, we compare the tendency of our method with that of an existing method.
We select \citet{wang-lu-2018-neural}'s method for comparison.\footnote{We do not use POS tags as one of input features for a fair comparison with our method.}
We train their model with the ACE-2005 dataset using their original implementation and repeat that 5 times.
The recall scores from the 1st level to the 4th level are $66.52\%$, $65.34\%$, $42.14\%$, and $50.00\%$, respectively.
The tendency about the difference across levels is common to \citet{wang-lu-2018-neural}'s method and our method, and the scores from our method (\tabref{tab:level-wise-recall}) are entirely higher than those from their method.
It is demonstrated that our method can extract both outer and inner entities better.
On the other hand, their method can extract crossing entities (two entities overlap but neither is contained in the other), although our method cannot.
Actually, their model outputs some crossing spans in our experiments.
In this case, we cannot analyze the results regarding precision scores in the same manner as \tabref{tab:level-wise-precision}.
There are cases where one cannot uniquely decide the level of an span nested within multiple crossing spans.
Regardless, our method cannot handle crossing entities.
However, crossing entities are very rare~\cite{lu-roth-2015-joint,wang-etal-2018-neural-transition}.
The test sets of ACE-2005 and GENIA have no crossing entities.
This property of our method does not have a negative impact the performance at least on the ACE-2005 and GENIA datasets.

\subsection{Error Analysis}

We manually scan the test set predictions on ACE-2005.
We find out that many of the errors can be classified into two types.

The first type is partial prediction error.
Given the following sentence: ``{\it Let me set aside the hypocrisy of a man who became president because of a lawsuit trying to eliminate everybody else's lawsuits, but instead focus on his own experience}''.
The annotation marks ``{\it a man who became president because of a lawsuit}'', but our model extracts a shorter or longer span. 
It is difficult to extract the proper spans of clauses that contain numerous modifiers.

The second type is error derived from pronominal mention.
Consider the following example: ``{\it They roar, they screech.}''.
These ``{\it They}''s refer to ``{\it tanks}'' in another sentence of the same document and are indeed annotated as VEH (Vehicle).
Our model fails to detect these pronominal mentions or wrongly labels them as PER (Person).
Document context should be taken into consideration to solve this problem.

These types of errors have been reported by \citet{katiyar-cardie-2018-nested,ju-etal-2018-neural,lin-etal-2019-sequence} and are still remaining as challenges.

\subsection{Running Time}

\begin{table}[t!]
\centering
\small
\resizebox{0.8\columnwidth}{!}{
\begin{tabular}{c|r}
  {\bf Maximal depth} & {\bf \# tokens per second} \\ \hline
  $1$ & 6,083 \\
  $2$ & 3,761 \\
  $3$ & 3,655 \\
  $4$ & 3,742 \\
  $5$ & 3,723 \\
  $\infty$ (no restriction) & 3,701 \\
\end{tabular}
}
\caption{Decoding speed on ACE-2005.}\label{tab:speed}
\end{table}

We investigate how our recursive decoding method impacts on the decoding speed in terms of the number of words processed per second.
We use the model trained with ACE-2005 used for \tabref{tab:level-wise-precision} and change the maximal depth of decoding to $1$, $2$, $3$, $4$, $5$, and $\infty$. 
When the maximal depth is $n$, our decoder Viterbi-decodes only from the 1st level to the $n$-th level.
Note that, when the maximal depth is $1$, the decoding process is completely the same as the Viterbi decoding of the standard CRF.
We run them on an Intel i7 (2.7 GHz) CPU.

Results are listed in \tabref{tab:speed}.
The processed words per second decrease by $38\%$ when the maximal depth varies from $1$ to $2$.
There are two main reasons for this phenomenon.
First, our decoder needs the processing for moving across different levels.
That processing is not necessary when the maximal depth is $1$.
Second, the number of the extracted spans at the 2nd level is large and not negligible ($12.5\%$ of that of the extracted spans at the 1st level as shown in \tabref{tab:level-wise-precision}).
The numbers of the extracted spans at the 3rd and lower levels are small, and then the processed words do not largely decrease when the maximal depth increases over $2$.
Regardless, our decoder does not take twice as long as the standard CRF on ACE-2005.

\begin{table}[t!]
\centering
\small
\resizebox{\columnwidth}{!}{
\begin{tabular}{l|lll}
  {\bf Method} &
  P ($\%$) & R ($\%$) & F1 ($\%$) \\
  \hline
  \citet{katiyar-cardie-2018-nested} &
  $72.3$ & $66.8$ & $69.7$ \\
  \citet{wang-etal-2018-neural-transition}$^\dag$\footnotemark &
  $74.9$ & $71.8$ & $73.3$ \\
  \citet{wang-lu-2018-neural}$^\dag$ &
  $78.0$ & $72.4$ & $75.1$ \\
  \citet{strakova-etal-2019-neural}$^\dag$\footnotemark &
  $78.92$ & $75.33$ & $77.08$ \\
  {\bf This work} &
  $79.93$ & $75.10$ & $\bm{77.44}$ \\ \hline
  \citet{strakova-etal-2019-neural} [BERT]$^\dag$ &
  $84.71$ & $83.96$ & $84.33$ \\
  {\bf This work} [BERT] &
  $85.23$ & $84.72$ & $\bm{84.97}$ \\ \hline
  \citet{strakova-etal-2019-neural} [BERT+FLAIR]$^\dag$ &
  $84.51$ & $84.29$ & $84.40$ \\
  {\bf This work} [BERT+FLAIR] &
  $85.94$ & $85.69$ & $\bm{85.82}$ \\
\end{tabular}
}
\caption{Comparison on ACE-2004. ``$\dag$'' indicates the methods using POS tags.}\label{tab:main-results-ace-2004}
\end{table}
\addtocounter{footnote}{-1}
\footnotetext{\citet{wang-etal-2018-neural-transition} did not report precision and recall scores. Instead of \citet{wang-etal-2018-neural-transition}, \citet{wang-lu-2018-neural} reported the scores for the model of \citet{wang-etal-2018-neural-transition}.}
\stepcounter{footnote}
\footnotetext{\citet{strakova-etal-2019-neural} did not report precision and recall scores in their paper. We asked the authors the scores, and they let us know.}

\subsection{Comparison on ACE-2004}

We compare our method with existing methods also on the ACE-2004 dataset.
We use the same splits as \citet{lu-roth-2015-joint}.
The setups are the same as those of our experiment on ACE-2005.
\tabref{tab:main-results-ace-2004} shows the results.
As shown, our method significantly outperforms existing methods.
Note that most of them use POS tags as an additional input feature whereas our method does not.

\subsection{Flat NER}

\begin{table}[t!]
\centering
\small
\resizebox{0.7\columnwidth}{!}{
\begin{tabular}{l|l}
  {\bf Method} &
  F1 ($\%$) \\
  \hline \hline
  \citet{wang-lu-2018-neural}$^\dag$ &
  $90.5$ \\
  \citet{strakova-etal-2019-neural}$^\dag$ &
  $90.77$ \\
  {\bf This work} &
  $91.14 \pm 0.04$ \\ \hline
  \citet{lample-etal-2016-neural}$^\ddag$  &
  $90.94$ \\
  \citet{ma-hovy-2016-end}$^\ddag$  &
  $91.21$ \\
  \citet{liu-etal-2019-gcdt}$^\ddag$  &
  $91.96 \pm 0.04$ \\
  {\bf This work} $-$ L\&D$^\ddag$ &
  $90.84 \pm 0.10$ \\ \hline \hline
  \citet{devlin-etal-2019-bert}$^\ddag$ &
  $92.80$ \\
  \citet{akbik-etal-2018-contextual}$^\ddag$ &
  $93.09 \pm 0.12$ \\
  \citet{liu-etal-2019-gcdt}$^\ddag$ &
  $93.47 \pm 0.03$ \\
  \citet{jiang-etal-2019-improved}$^\ddag$ &
  $93.47$ \\
  \citet{baevski-etal-2019-cloze}$^\ddag$ &
  $93.5$ \\
\end{tabular}
}
\caption{Comparison on CoNLL-2003. We group methods into two types. The first group consists of the methods that do not use any contextual word embeddings. The second one consists of the methods that use contextual word embeddings such as BERT and FLAIR. ``$\dag$'' indicates the methods using POS tags. ``$\ddag$'' indicates the methods not designed to extract nested entities.}\label{tab:main-results-conll-2003}
\end{table}

To assess how our model works on flat NER task, we additionally evaluate our model on CoNLL-2003 \cite{tjong-kim-sang-de-meulder-2003-introduction}, which are annotated with outermost entities only.
The setups here are the same as those of our experiment on ACE-2005.
We not only prepare our proposed model but also the ablated model without our training nor decoding method, as in Section \ref{ssec:ablation}.
The former model can extract spans nested within other extracted spans regardless of the property of the dataset, but the latter model never extracts spans within other extracted spans.
We use the 100-dimensional GloVe embeddings for both models as in our previous experiments.

The results are in \tabref{tab:main-results-conll-2003}.
We compare our method with existing methods that do not adopt any contextual word embeddings (the upside of \tabref{tab:main-results-conll-2003}) here, although we also show results from recent work with contextual word embeddings for reference.
First, in comparison with the methods designed for nested NER~\cite{wang-lu-2018-neural,strakova-etal-2019-neural}, our method performs better even on CoNLL-2003.
This means that our method works well on not only nested NER but also flat NER.
Next, we compare to methods that can handle only flat NER.
\tabref{tab:main-results-conll-2003} shows that our method is comparable to the standard BiLSTM-CRF models~\cite{lample-etal-2016-neural,ma-hovy-2016-end} on CoNLL-2003.
However, note that there are some differences between the experiments of the previous studies~\cite{lample-etal-2016-neural,ma-hovy-2016-end} and our experiment.
For example, different word embeddings are used, or the hidden size of LSTM is not aligned.
Nevertheless, we can compare our proposed model to the ablated model.
As shown in \tabref{tab:main-results-conll-2003}, there is a significant gap ($p < 0.005$ with the permutation test) between the two scores, $91.14 (\pm 0.04) \%$ and $90.84 (\pm 0.10) \%$.
We analyze this gap in detail and then find out that our proposed model performs well especially in the cases where it is difficult to decide which is suitable, an inner span or an outer span.
Given the following sentence: ``{\it An assessment group made up of the State Council 's Port Office , the Civil Aviation Administration of China , the General Administration of Customs and other authorities had granted the airport permission to handle foreign aircraft , Xinhua said .}''.
In the CoNLL-2003 dataset, the four spans ``{\it State Council}'', ``{\it Civil Aviation Administration of China}'', ``{\it General Administration of Customs}'', and ``{\it Xinhua}'' are annotated as ORG (Organization).
The both models correctly detect the latter three entities in most trials, but the ablated model tends to extract ``{\it State Council 's Port Office}'' instead of ``{\it State Council}''.
On the other hand, our proposed model tends to extract both ``{\it State Council 's Port Office}'' and ``{\it State Council}''.
``{\it State Council 's Port Office}'' is indeed a false-positive, but our model can detect the correct entity span ``{\it State Council}'' more steadily than the ablated model.
Thus, our proposed model achieves the higher F1-score.

Recently, \citet{liu-etal-2019-gcdt} proposed a new architecture for sequence labeling, which can capture global information at the sentence level better than BiLSTM, and reported an F1-score of $91.96\%$ when using conventional word embeddings ($93.47\%$ when using BERT).
It is true that our model based on BiLSTM does not perform as well as their model, but our decoder can be combined with their proposed encoder.
We leave it for future work.

\section{Related Work}

\citet{alex-etal-2007-recognising} proposed several ways to combine multiple CRFs for such tasks.
They found out that, when they cascaded separate CRFs of each entity type by using the output from the previous CRF as the input features of the current CRF, best performance was yielded.
However, their method could not handle nested entities of the same entity type.
In contrast, \citet{ju-etal-2018-neural} dynamically stacked multiple layers that recognize entities sequentially from innermost ones to outermost ones.
Their method can deal with nested entities of the same entity type.

\citet{finkel-manning-2009-nested} proposed a CRF-based constituency parser for this task such that
each named entity is a node in the parse tree.
However, its time complexity is the cube of the length of a given sentence, making it not scalable to large datasets involving long sentences.
Later on, \citet{wang-etal-2018-neural-transition} proposed a scalable transition-based approach, a constituency forest (a collection of constituency trees).
Its time complexity is linear in the sentence length.

\citet{lu-roth-2015-joint} introduced a mention hypergraph representation for capturing nested entities as well as crossing entities (two entities overlap but neither is contained in the other).
One issue in their approach is the spurious structures of the representation.
\citet{muis-lu-2017-labeling} incorporated mention separators to address the spurious structures issue, but it still suffers from the structural ambiguity issue.
\citet{wang-lu-2018-neural} proposed a hypergraph representation free of structural ambiguity.
However, they introduced a hyperparameter, the maximal length of an entity, to reduce the time complexity.
Setting the hyperparameter to a small number results in speeding up but ignoring longer entity segments.

\citet{katiyar-cardie-2018-nested} proposed another hypergraph-based approach that learns the structure using an LSTM network in a greedy manner.
However, their method has a hyperparameter that sets a threshold for selecting multiple candidate mentions.
It must be carefully tuned for adjusting the trade-off between recall and precision.

\citet{sohrab-miwa-2018-deep} proposed a neural exhaustive model that enumerates all possible spans as potential entity mentions and classifies them.
However, they also use the maximal-length hyperparameter to reduce time complexity.

\citet{fisher-vlachos-2019-merge} proposed a novel neural network architecture that merges tokens or entities into entities forming nested structures and then labels each of them.
Their architecture, however, needs the maximal nesting level hyperparameter.
\citet{lin-etal-2019-sequence} proposed a sequence-to-nuggets architecture that first identify anchor words of all mentions and then recognize the mention boundaries for each anchor word.
Their method also use the maximal-length hyperparameter to reduce time complexity.

\citet{strakova-etal-2019-neural} proposed an encoding algorithm to allow the modeling of multiple named entity labels in a linearized scheme and proposed a neural model that predicts sequential labels for each token.
\citet{zheng-etal-2019-boundary} proposed a method that can detect entities boundaries with sequence labeling models.
These two methods do not require special hyperparameters.
They can also deal with crossing entities as well as nested entities in contrast to our method, but our experiments demonstrate that our method can perform well because crossing entities are very rare~\cite{lu-roth-2015-joint,wang-etal-2018-neural-transition}.

\section{Conclusion}

We propose learning and decoding methods for extracting nested entities.
Our decoding method iteratively recognizes entities from outermost ones to inner ones in an outside-to-inside way.
It recursively searches a span of each extracted entity for nested entities with second-best sequence decoding.
We also design an objective function for training that ensures our decoding algorithm.
Our method has no hyperparameters beyond those of conventional CRF-based models.
Our method achieves $85.82\%$, $84.34\%$, and $77.36\%$ F1-scores on ACE-2004, ACE-2005, and GENIA datasets, respectively.

For future work, one interesting direction is joint modeling of NER with entity linking or coreference resolution.
Previous studies \cite{durrett-klein-2014-joint,luo-etal-2015-joint,nguyen-etal-2016-j,martins-etal-2019-joint} demonstrated that leveraging mutual dependency of the NER, linking, and coreference tasks could boost each performance.
We would like to address this issue while taking nested entities into account.

\section*{Acknowledgements}
We thank Aldrian Obaja Muis for helpful comments, and many anonymous reviewers and the action editor for helpful feedback on various drafts of the paper. We are also grateful to Jana Strakov{\'a} for sharing experimental results. Eduard Hovy was supported in part by DARPA grant FA8750-18-2-0018 funded under the AIDA program.

\bibliography{tacl2018}
\bibliographystyle{acl_natbib}

\end{document}